\documentclass[conference]{ieeetran}
\IEEEoverridecommandlockouts                   
\usepackage{graphicx}
\usepackage[T1]{fontenc}
\usepackage[utf8]{inputenc}
\usepackage{cite}
\usepackage[plainpages=false,hypertexnames=false,pdfpagelabels]{hyperref}

\usepackage[fleqn]{mathtools}
\usepackage{bm}
\usepackage{amssymb}

\usepackage[labelfont=bf,labelsep=period]{caption}
\usepackage[]{subcaption}
\usepackage[fleqn]{mathtools}
\usepackage{bm}
\usepackage{amssymb}
\usepackage{floatrow}
\usepackage{tabularx}
\usepackage{pdfcomment}
\usepackage[dvipsnames]{xcolor}

\usepackage{array}
\newcolumntype{?}{!{\vrule width 1pt}}

\defineavatar{tw}{author=Yoyek,color=yellow}

\newfloatcommand{capbtabbox}{table}[][\FBwidth]

\usepackage{listings}
\newcommand{\Fig}[1]{Fig.~{\ref{#1}}}

\def\tiago{\texttt{TIAGo}}
\def\ros{\texttt{ROS}}

\newcommand{\eaco}[1]{\mathit{#1/aco}}
\newcommand{\eapl}{\mathit{/apl}}
\newcommand{\eatr}[1]{\mathit{#1/atr}}
\newcommand{\eath}[1]{\mathit{#1/ath}}
\newcommand{\ears}{\mathit{/ars}}
\newcommand{\eady}[1]{\mathit{#1/ady}}

\begin{document}

	\title{An intent-based approach for creating assistive robots' control systems
	\thanks{This work was funded within the INCARE AAL-2017-059 project ,,Integrated Solution for Innovative Elderly Care'' by the AAL JP and co-funded by the AAL JP countries (National Centre for Research and Development, Poland under Grant AAL2/2/INCARE/2018). The work was initially submitted to the 25th International Conference on Methods and Models in Automation and Robotics (MMAR) that was cancelled due to COVID-2019. The authors would like to thank the MMAR reviewers for their effort to help to improve the paper and Maciej Bogusz for his work on robot database development.}
	}


	
	%
	%
	
	\author{\IEEEauthorblockN{Tomasz Winiarski}
	\IEEEauthorblockA{
		Warsaw University of Technology,\\
		Institute of Control and\\
		Computation Engineering\\
		Warsaw, Poland,\\
		Email: tomasz.winiarski@pw.edu.pl}
	\and
	\IEEEauthorblockN{Wojciech Dudek}
	\IEEEauthorblockA{
		Warsaw University of Technology,\\
		Institute of Control and\\
		Computation Engineering\\
		Warsaw, Poland}
	\and
	\IEEEauthorblockN{Maciej Stefańczyk}
	\IEEEauthorblockA{
		Warsaw University of Technology,\\
		Institute of Control and\\
		Computation Engineering\\
		Warsaw, Poland}
	\and
	\IEEEauthorblockN{Łukasz Zieliński}
	\IEEEauthorblockA{
		Warsaw University of Technology,\\
		Institute of Control and\\
		Computation Engineering\\
		Warsaw, Poland}
	\and
	\IEEEauthorblockN{Daniel Giełdowski}
	\IEEEauthorblockA{
		Warsaw University of Technology,\\
		Institute of Control and\\
		Computation Engineering\\
	Warsaw, Poland}
	\and
	\IEEEauthorblockN{Dawid Seredyński}
	\IEEEauthorblockA{
		Warsaw University of Technology,\\
		Institute of Control and\\
		Computation Engineering\\
	Warsaw, Poland}
	}

	\maketitle
	\begin{abstract}	
		The current research standards in robotics demand general approaches to robots' controllers development. In the assistive robotics domain, the human-machine interaction plays a~substantial role. Especially, the humans generate intents that affect robot control system. In the article an approach is presented for creating control systems for assistive robots, which reacts to users' intents delivered by voice commands, buttons, or an operator console. The whole approach was applied to the real system consisting of customised \tiago{} robot and additional hardware components. The exemplary experiments performed on the platform illustrate the motivation for diversification of human-machine interfaces in assistive robots.
	\end{abstract}


\section{Introduction}
\label{sec:introduction}

The current demographical processes demand new technologies to support a~large number of Elderly or disabled persons. Assistive and social robots respond to it \cite{leite2013social,martinez2020socially}. Although some aspects of assistive robot development have a~general nature (e.g. low-level motor control), others are more specific. Human--machine interfaces (HMIs) play a~vital role in practical applications of assistive robots. Human--machine interaction is being investigated both for mobile robots \cite{kkedzierski2015design} and stationary HMI devices \cite{dziergwa2018long}. One of the main problems encountered is to maintain reliability of communication.            

The goal of our work is to at first determine appropriate HMIs to provide robust multimodal interaction of a~human and robot-based system. We concentrate on the armless systems, as they are expected to dominate the market, because of the relatively low price. Efficient research demands frameworks~\cite{harris2011survey} that reduce the time to develop software by code reuse and utilisation of supporting tools. Hence, we propose an intent based general approach to assistive robot's controller development that takes into account the current technologies that support the efficient usage of the previously chosen HMIs. 

The research process started with the determination of the HMIs based on the analysis of the existing robotic platforms (sec.~\ref{sec:survey}).
The general structure of the based system is depicted in sec.~\ref{sec:structure}, while the main aspects of its behaviour in sec.~\ref{sec:behaviour}. Sec.~\ref{sec:evaluation} presents a~practical application of our approach, i.e. the system build basing on \tiago{} robot. The exemplary experiments illustrate the importance of human-machine interface multimodality. The paper is concluded in sec.~\ref{sec:conclusions}.

\section{Survey of robotic platforms and their HMIs}
\label{sec:survey}

A~review of 58 different robotic platforms was conducted. The functions of examined robots varied from entertainment platforms, through assistive services, to helping people with day-to-day activities and supporting Elderly (Tab.~\ref{tab:review}).

\begin{table*}[t]
	\footnotesize 
	\begin{tabular}[b]{|l|l|l|p{3mm}|p{3mm}|p{3mm}|p{3mm}??l|l|l|p{3mm}|p{3mm}|p{3mm}|p{3mm}|}
		\hline
		No. & Robot name & Producer & \parbox[t]{2mm}{\rotatebox[origin=r]{90}{ Interactive tablet}}
		& \parbox[t]{2mm}{\rotatebox[origin=r]{90}{ Text to speech}}
		& \parbox[t]{2mm}{\rotatebox[origin=r]{90}{ Voice recognition}}
		& \parbox[t]{2mm}{\rotatebox[origin=r]{90}{ Physical buttons*}}&
		No. & Robot name & Producer & \parbox[t]{2mm}{\rotatebox[origin=r]{90}{ Interactive tablet}}
		& \parbox[t]{2mm}{\rotatebox[origin=r]{90}{ Text to speech}}
		& \parbox[t]{2mm}{\rotatebox[origin=r]{90}{ Voice recognition}}
		& \parbox[t]{2mm}{\rotatebox[origin=r]{90}{ Physical buttons*}}\\\hline
		
		1&Oro&ROBOT-ERA&\checkmark&\checkmark&?&R&
		31&HSR&Toyota&X&\checkmark&\checkmark&X\\\hline
		
		2&Coro&ROBOT-ERA&\checkmark&\checkmark&\checkmark&X&
		32&MARIO&MARIO Project&\checkmark&\checkmark&X&X\\\hline
		
		3&Doro&ROBOT-ERA&\checkmark&\checkmark&\checkmark&X&
		33&VirtualME&CrossWing Inc&\checkmark&X&X&X\\\hline
		
		4&MAGGIE&Robotics Lab - &\checkmark&\checkmark&\checkmark&X&
		34&AMIGO&Eindhoven &X&\checkmark&\checkmark&X\\
		&&Universidad Carlos &&&&&&&University &&&&\\
		&&III de Madrid&&&&&&&of Technology&&&&\\\hline
		
		5&Ferry&Paaila&?&X&X&X&
		35&AMY A1&AMY Robotics&\checkmark&\checkmark&\checkmark&X\\\hline
		
		6&Pati&Paaila&\checkmark&\checkmark&\checkmark&R&
		36&ARMadillo&RoboTiCan&\checkmark&X&X&X\\\hline
		
		7&Ginger&Paaila&X&\checkmark&\checkmark&X&
		37&Fetch&Fetch Robotics&X&X&X&X\\\hline
		
		8&Care-O-bot 3&Fraunhofer IPA&\checkmark&\checkmark&\checkmark&X&
		38&ForteRC&Ingeniarius, Ltd.&\checkmark&X&X&X\\\hline
		
		9&Care-O-bot 4&Fraunhofer IPA&\checkmark&X&\checkmark&X&
		39&Sunbot-II&Siasun&X&\checkmark&\checkmark&X\\\hline
		
		10&Relay&Savioke&\checkmark&X&X&X&
		40&Sunbot-III&Siasun&\checkmark&\checkmark&\checkmark&X\\\hline
		
		11&Stevie&Trinity College&X&\checkmark&?&X&
		41&RB-1&Robotnik &X&X&X&X\\
		&&Dublin&&&&&&&Automation S.L.L.&&&&\\\hline
		
		12&HOBBIT&Hobbit&\checkmark&\checkmark&X&X&
		42&RP-VITA&CORDAMED&\checkmark&X&\checkmark&X\\\hline
		
		13&Zenbo&ASUS&\checkmark&\checkmark&\checkmark&X&
		43&i-foot&Toyota&X&X&X&R\\\hline
		
		14&I-do&KUKA&X&X&X&X&
		44&TUG Exchange&Aethon&X&\checkmark&X&X\\\hline
		
		15&Aibo&Sony&X&X&\checkmark&X&
		45&TUG T3&Aethon&X&\checkmark&X&X\\\hline
		
		16&CHiP&WowWee&X&X&\checkmark&X&
		46&TUG Drawer&Aethon&X&\checkmark&X&\checkmark\\\hline
		
		17&Arti&Warsaw University&X&?&?&X&
		47&TUG for&Aethon&\checkmark&\checkmark&X&X\\
		&&of Technology&&&&&&Hospitality&&&&&\\\hline
		
		18&Temi&Temi&\checkmark&\checkmark&\checkmark&X&
		48&K5&Knightscope&X&\checkmark&X&X\\\hline
		
		19&Robelf&Robelf&\checkmark&\checkmark&\checkmark&?&
		49&ReMeDi&ReMeDi consortium &X&X&X&R\\\hline
		
		20&Honeybot&Honeybot&\checkmark&\checkmark&X&R&
		50&Hugo/GrowMU&GrowMeUp&\checkmark&\checkmark&\checkmark&X\\\hline
		
		21&Kuri&Mayfield Robotics&X&X&\checkmark&X&
		51&MOVO&Kinova&X&X&X&X\\\hline
		
		22&Robot Prison &Kyonggi &\checkmark&X&X&X&
		52&CASERO&MLR System &\checkmark&X&X&R\\
		&Guard&University&&&&&&&GmbH&&&&\\\hline
		
		23&Sanbot Elf&Sanbot Innovation&\checkmark&\checkmark&\checkmark&X&
		53&RAMCIP&CERTH&\checkmark&\checkmark&\checkmark&X\\\hline
		
		24&Robina&Toyota&X&\checkmark&\checkmark&X&
		54&Pepper&SoftBank Robotics&\checkmark&\checkmark&\checkmark&X\\\hline
		
		25&ROBEAR&RIKEN&?&X&X&X&
		55&Robovie II&ATR&X&\checkmark&\checkmark&X\\\hline
		
		26&Sunbot-I&Siasun&X&\checkmark&\checkmark&X&
		56&PeopleBot&Adept&X&\checkmark&\checkmark&X\\\hline
		
		27&Tiro Robot&Hanool Robotics&\checkmark&\checkmark&\checkmark&X&
		57&FLASH/&Wrocław University &X&\checkmark&\checkmark&X\\
		&&Corp&&&&&&FLASH MK II&of Science and &&&&\\
		&&&&&&&&&Technology&&&&\\\hline
		
		28&Benebot&ECOVACS&X&\checkmark&\checkmark&R&
		58&LoweBot&Lowe's Innovation &\checkmark&X&\checkmark&X\\
		&&ROBOTICS&&&&&&&labs&&&&\\\hline
		
		29&REEM&PAL Robotics&\checkmark&\checkmark&X&X&
		&&&&&&\\\hline
		
		30&TIAGo&PAL Robotics&X&\checkmark&\checkmark&X&
		&&&&&&\\\hline
		
	\end{tabular}\\
	\small
	* Emergency stop buttons are not regarded,
	\checkmark -- Robot possesses feature, 
	X -- Robot does not possess feature, 
	? -- The possession of\\ feature is suggested but could not be confirmed, 
	R -- Access to the physical button is restricted because of its location.
	\caption{Review of selected functions for 58 different robotic platforms}
	\label{tab:review}
\end{table*}

The study inspected the robots for the chosen features:
\begin{enumerate}
	\item possession of the built in, interactive, touch-screen tablet,
	\item ability to speak,
	\item ability to recognise voice commands,
	\item possession of physical buttons and their placement.
\end{enumerate}
Out of 58 examined robotic platforms:
\begin{itemize}
	\item 30 possessed the specified tablet,
	\item 36 were able to speak,
	\item 32 were capable of voice commands recognition,
	\item 8 possessed some physical buttons (emergency buttons are not counted, because they are not the regular communication medium) and only one robot had buttons that could be accessed from all sides of the robot.
\end{itemize}
Note: in some cases, some of the robot features were not described. Such robots were not included in the above juxtaposition.

The above survey visualises that voice communication and tablets are the main HMIs, while the buttons are absent or used for emergency procedures only. To obtain the highest reliability of voice communication, nowadays the cloud-based services can be used. Although the usage of tablets touch-screen, as the different communication interface to voice, is versatile, we opt for buttons. In comparison to buttons, tablet is bigger, heavier, prone to physical damage. It needs to be charged often, its usage is more complicated, especially for Elderly~\cite{barnard2013learning}.
 The number of assistive robots applications can be handled with buttons, in order to, e.g., call the robot (the appropriate button is located in the home environment), acknowledge the command spoken by the robot, etc. The buttons should be mounted in the proper places to be accessed easily and intuitively.
For the convenience of research, development, maintenance or even regular operation, an operator/caregiver console should be included in the system.

\vspace{2cm}

\section{General Structure and Behaviour of a~system}
\label{sec:structure-behaviour}

Our approach to control system creation is based on the RAPP architecture~\cite{zielinski2017variable} extended by the Agent types introduced in~\cite{Dudek-multitasking-romoco-2019} -- a~Task Harmoniser Agent and  a~Task Requester Agent. It is described with usage of EARL \cite{earl2020}  -- Embodied Agent-based\cite{kornuta-bpan-2020} cybeR-physical control systems modelling Language (version~1.2)~\cite{earl12}. 

\subsection{Structure}	
\label{sec:structure}

\begin{figure}[htt]
	\includegraphics[width=\linewidth]{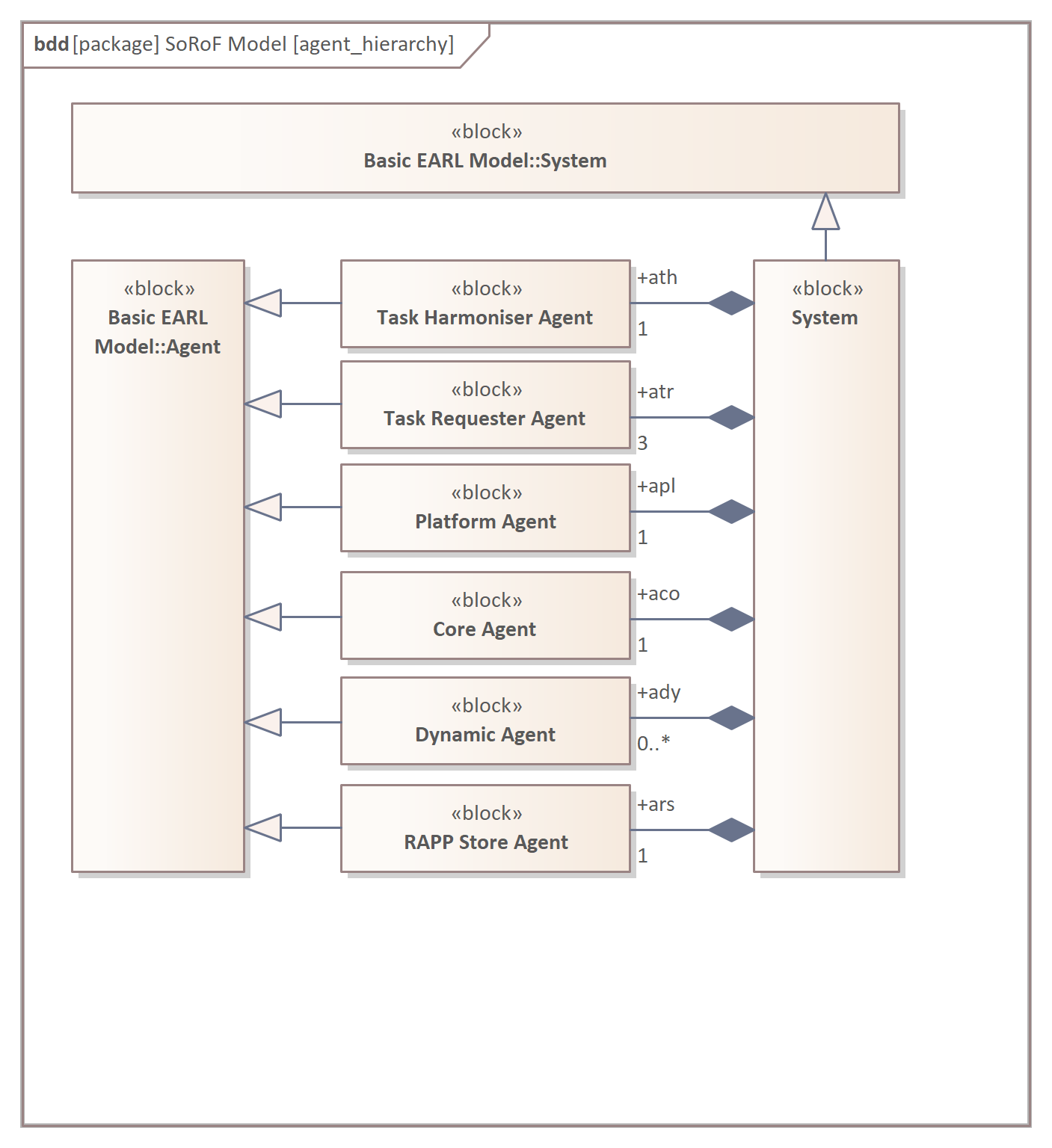}
	\caption{Agent hierarchy}
	\label{fig:agent_hierarchy}
\end{figure}

The whole System is decomposed into various types of Agents derived from the general Agent <<block>> (\Fig{fig:agent_hierarchy}). 

\begin{figure}[htb]
	\centering
	\includegraphics[width=.9\linewidth]{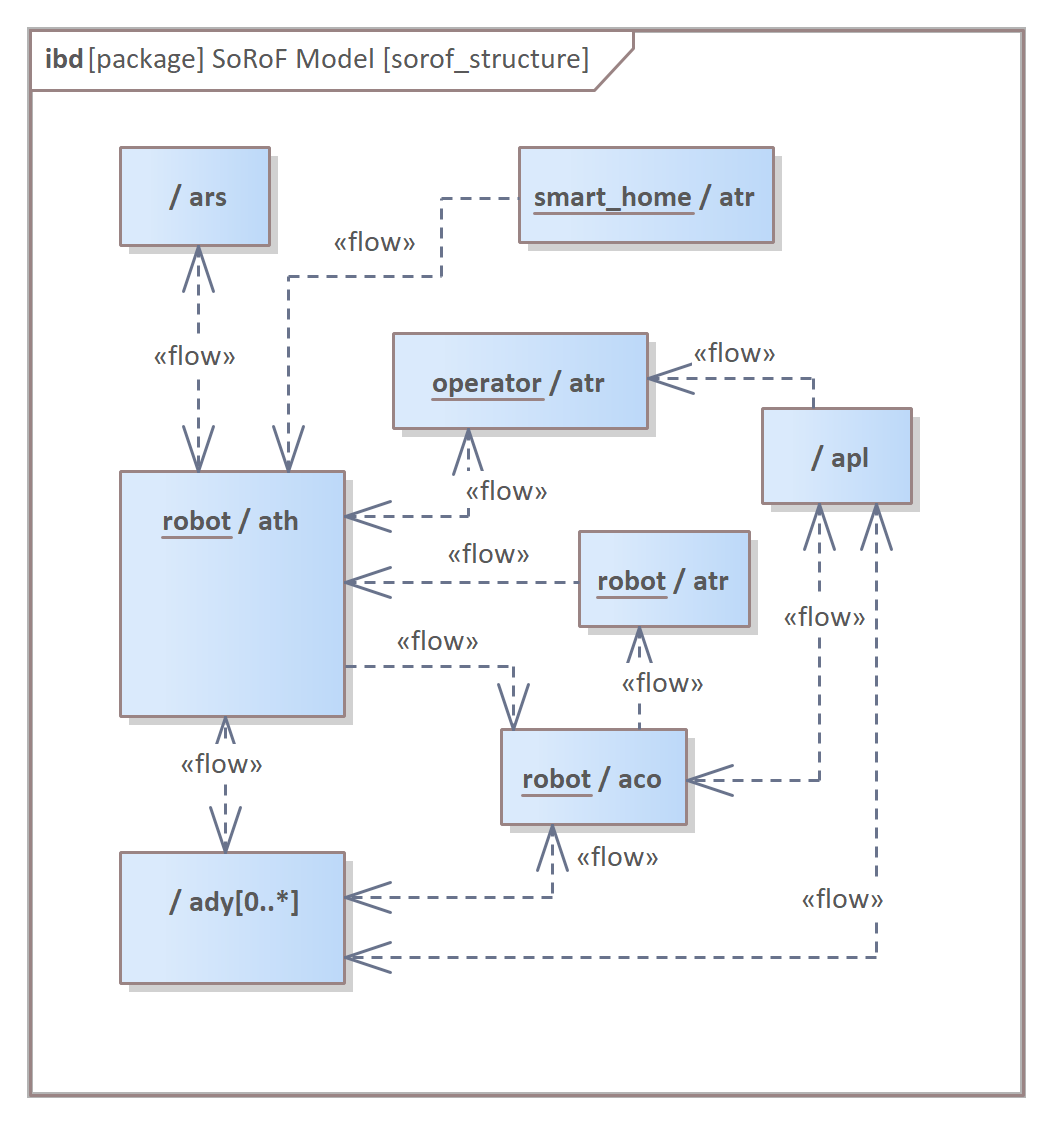}
	\caption{Inter Agent communication in the System}
	\label{fig:sorof-agent-communication}
\end{figure}

The Agents instances are the parts of the System with particular names. The permissible communication in the System is shown in internal block diagram (\Fig{fig:sorof-agent-communication}).

\subsubsection{Core Agent} drives the hardware of the robot and provides the services to be called by other Agents:
\begin{itemize}
	\item navigation configuration and goal set,
	\item control of neck and torso lift,
	\item text-to-speech,
	\item translation of a~voice command into an intent.
\end{itemize}

In the presented System there is one robot operating in the environment, so there is only one Core Agent named $\eaco{robot}$.

\subsubsection{Dynamic Agents} are parts of the System, which utilise services delivered by other Agents in order to perform a~task that a~particular Dynamic Agent is designed for. A~Dynamic Agent interprets data possessed from other Agents and compute control for the robot, guiding it to perform the task implemented in the Dynamic Agent. An exemplary specification of a~Dynamic Agent was described in~\cite{mmar_dudek_distributed-2016}. There can be multiple Dynamic Agents in the System, however, at most one can be initialised at a~time.

\subsubsection{Platform Agent} aggregates System-wide computational services. As the Platform Agent from its definition operates in the cloud~\cite{dudek:2017-jamris-cloud-computing} and has huge computation power and storage space, most of the behaviours requiring high computation power are delegated here. However, some of the computationally complex behaviours have to be done by Agents located in the robot computer as they are latency-sensitive or process sensitive personal data. The method of a~System behaviours delegation between Agents was presented in~\cite{dudek:2016-automation}. There is only one Platform Agent in the System.

\subsubsection{Task Requester Agents} are parts of the System, which basing on the data from robot's or additional sensors request the Task Harmoniser Agent to perform tasks. There are three Task Requester Agents in the System: $\eatr{smart\_home}$, $\eatr{robot}$ and $\eatr{operator}$. The first one is an interface for a~smart home platform. It requests tasks basing on the defined rules. The $\eatr{robot}$ uses user intents captured by the robot hardware (e.g. microphone), and the $\eatr{operator}$ uses an application that is used for robot supervision.


\subsubsection{Task Harmoniser Agent} receives task requests from the Task Requester Agents and schedules Dynamic Agents basing on a~specified algorithm (e.g. priorities). The problem of Dynamic Agents scheduling was introduced in \cite{Dudek-multitasking-romoco-2019}. It communicates with RAPP Store Agent in order to download a~proper task files implementing a~Dynamic Agent. There is one Task Harmoniser Agent in the System named $\eath{robot}$.

\subsubsection{RAPP Store Agent} keeps files with executable code that implement various tasks available for a~System and provides a~service allowing to download them.

\subsection{Behaviour}
\label{sec:behaviour}

\begin{figure}[htb]
	\includegraphics[width=.7\linewidth]{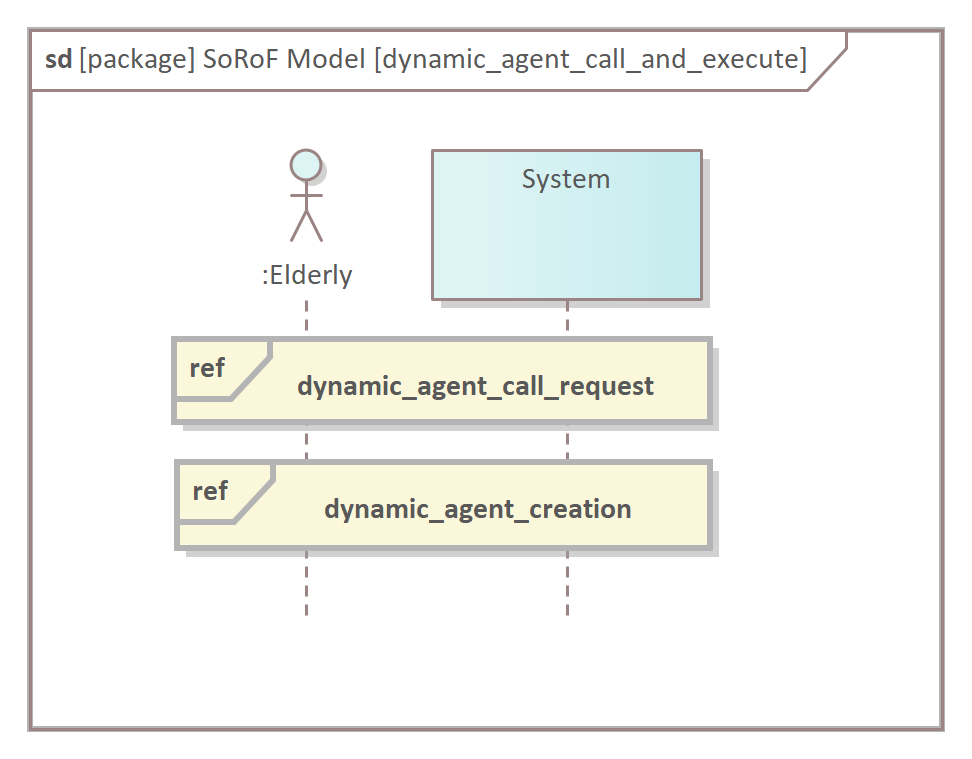}
	\caption{Dynamic Agent call request and creation}
	\label{fig:voice-commands}
\end{figure}

\begin{figure}[htb]
	\includegraphics[width=\linewidth]{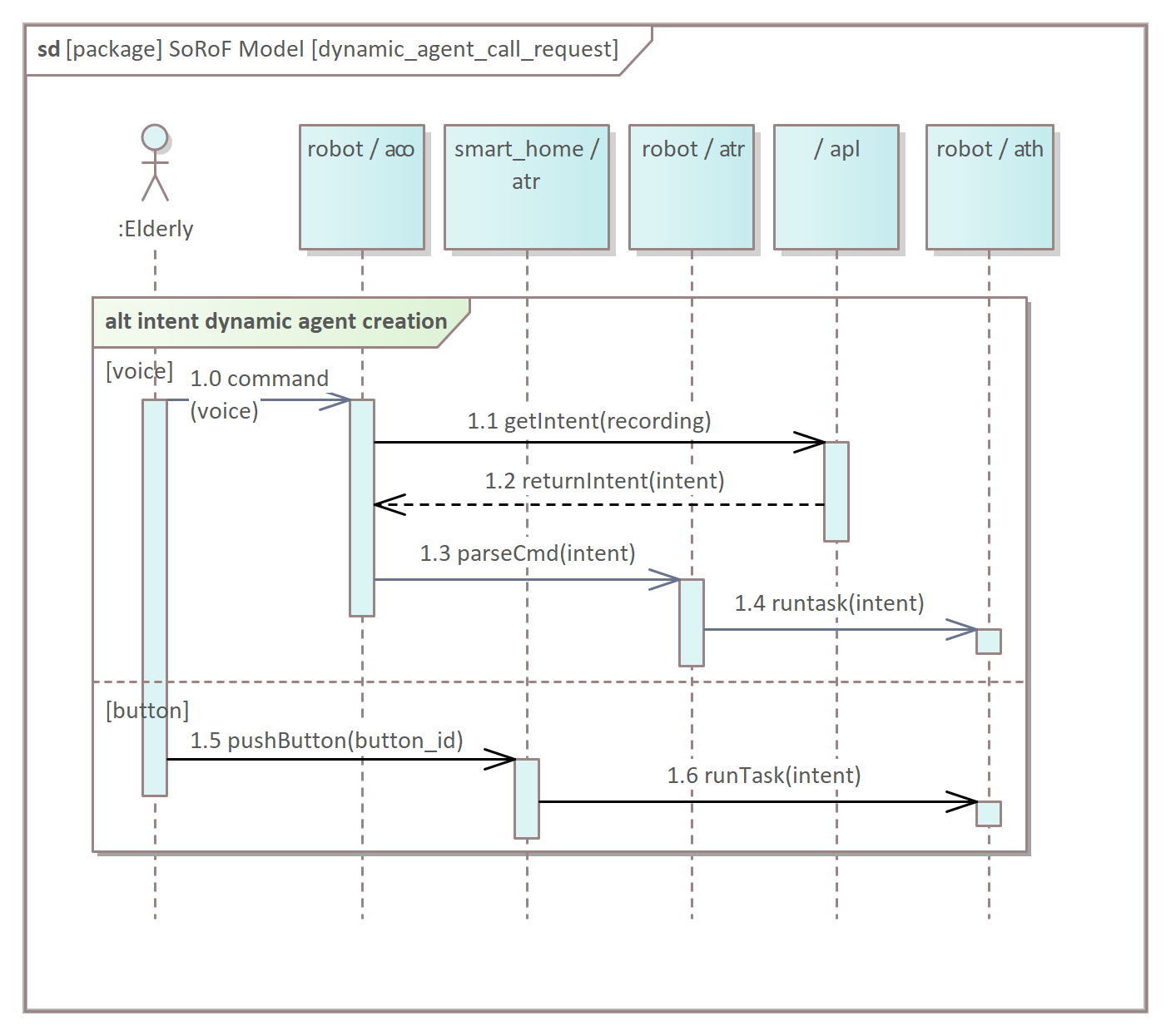}
	\caption{Dynamic Agent call request}
	\label{fig:da-call-request}
\end{figure}

One of the main aspects of communication between Agents is shown in Fig.~\ref{fig:voice-commands}.
At first, an Elderly commands the robot either with voice or with a~button (Fig.~\ref{fig:da-call-request}).
In the first case, the voice is recorded by $\eaco{robot}$ and the recording
is sent to $\eapl$. The Platform Agent runs speech-to-text algorithms and sends back the detected intent
to $\eaco{robot}$. An intent represents an interpretation of the voice command.
Next, the intent is sent to $\eath{robot}$ via $\eatr{robot}$.
In the second case, a~button is triggered, and $\eatr{smart\_home}$ sends to $\eath{robot}$ an intent related to the button.
At this point there are two alternatives (Fig.~\ref{fig:da-creation}):
\begin{figure}[htb]
	\includegraphics[width=\linewidth]{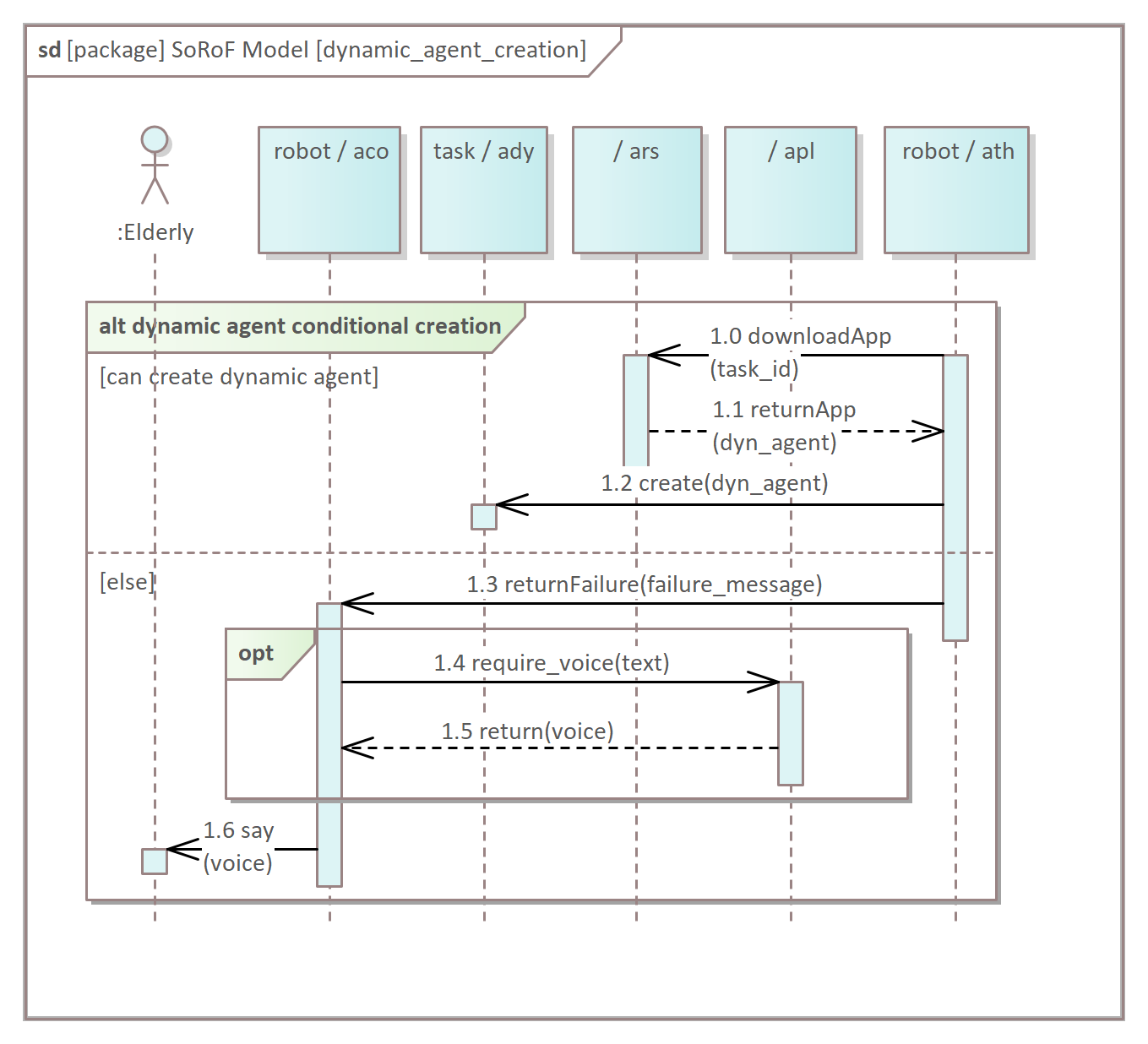}
	\caption{Dynamic Agent creation}
	\label{fig:da-creation}
\end{figure}
\begin{itemize}
	\item $\eady{task}$ can be created, so task harmoniser $\eath{robot}$ downloads the code of Dynamic Agent $\eady{task}$ from the RAPP Store Agent $\ears$. It is done by sending a~message
	from $\eath{robot}$ to $\ears$. In response, a~message
	containing the code of $\eady{task}$ is sent in opposite direction and $\eath{robot}$ creates $\eady{task}$,
	\item $\eady{task}$ cannot be created, so $\eath{robot}$ sends a~message
	with human-readable description of failure reason to $\eaco{robot}$. If needed, the Core Agent requests text-to-speech from $\eapl{}$ for the failure reason text. The received voice response is played back to human.
\end{itemize}
The possible causes of $\eady{task}$ creation failure are:
\begin{itemize}
	\item another Dynamic Agent is already running, and it has higher priority than the requested one,
	\item the detected intent is not a~command for creation of a~new Dynamic Agent.
\end{itemize}
In order to avoid repeating text-to-speech $\eapl{}$ queries, the recordings are cached by $\eaco{robot}$ and they can be reused (`opt' frame in Fig.~\ref{fig:da-creation}).

\begin{figure}[htb]
	\includegraphics[width=\linewidth]{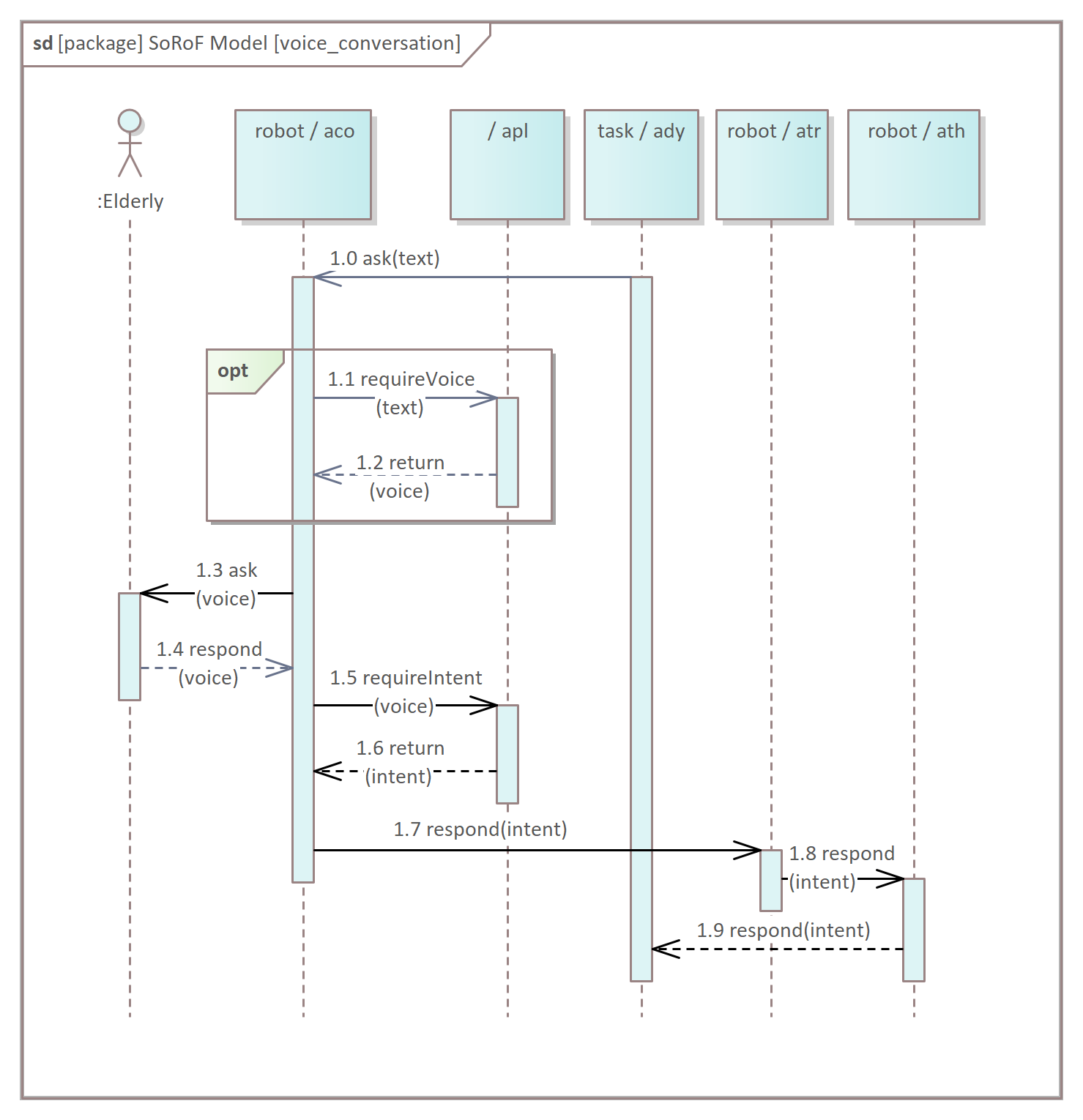}
	\caption{Voice conversation}
	\label{fig:voice_conversation}
\end{figure}

Another example of communication between agents is shown in Fig.~\ref{fig:voice_conversation}.
In this case, the dynamic agent $\eady{task}$ initiates a~conversation with human by sending a~message
to $\eaco{robot}$. Text-to-speech, if needed, is realized in $\eapl$, thus $\eaco{robot}$ sends message with text
to $\eapl$ and receives back a~message
that contains a~wave file. The voice is played to human. 
The human says a~response, and the recorded voice is processed by $\eapl$ to detect intent, similarly as in the previous example. The intent is then passed through $\eatr{robot}$ and $\eath{robot}$ to $\eady{task}$.


\section{Evaluation}
\label{sec:evaluation}

\subsection{Experimental platform}

\begin{figure}[htb]
	\includegraphics[width=\columnwidth]{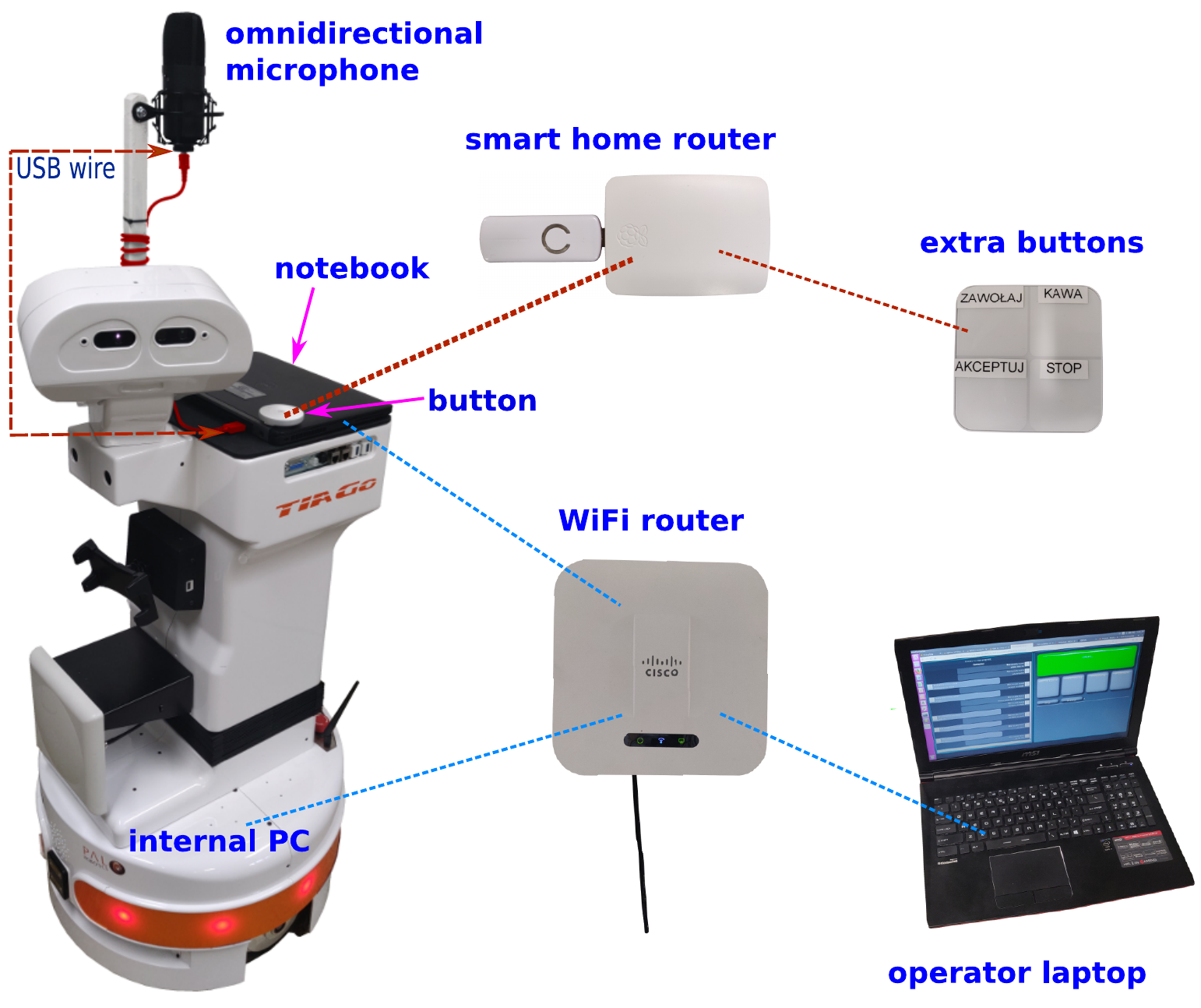}
	\caption{Customised \tiago{} robot and other hardware components involved in practical application}
	\label{fig:rico-mico}
\end{figure}

The experimental platform is presented in Fig.~\ref{fig:rico-mico}. 
\tiago{} robot was employed, as it is a~robust, \ros{}-based\cite{quigley2009ros} mobile platform used in modern research projects, e.g.~\cite{mocanu2018human}, especially when assistive robotics is concerned~\cite{PiasekTiago}.
An important reason to chose \tiago{} was the compatibility of Embodied Agent theory with mobile bases utilising \ros{} (e.g., a~robot with various modes of locomotion\cite{mmar_seredynski_control-2016}).
 A~detailed description of the \tiago{} platform and its capabilities are presented in~\cite{pages2016tiago}. The robot with its internal PC and all software provided by PAL is specified as a~single embodied Agent $\eaco{robot}$ (Fig.~\ref{fig:rico-mico}). 
Due to $\eaco{robot}$ internal PC performance limits, an additional notebook was required to perform voice acquisition and high-level task execution. Thus, a~part of  $\eaco{robot}$ Agent responsible for voice processing, and Agents $\eatr{robot}$, $\eath{robot}$ and all $\eady{}$ are realised as software (\ros{} nodes) running on the notebook.


The notebook and the internal PC of \tiago{} communicate with each other through wireless network. \ros{} was used as the main implementation platform.
Agent $\eatr{smart\_home}$ is composed of devices of smart home that are connected through wireless network.
Dynamic Agents $\eady{}$ are implemented using SMACH~\cite{Boren:2010}, however, any other framework compatible with \ros{} can be used.

\subsection{Exemplary experiments}

During the initial tests of the system, some audio capture problems were identified. \tiago{} is
equipped with a~built-in microphone \cite{grama2018adding}, but it is mounted inside the robot's
body, close to the head's motor. Stiff mounting of the microphone causes a~high level of rumble noises 
during the movement, which made it almost impossible to recognise speech. To counteract this problem
mounting was changed to soft sponge, but limited space inside the robot made it hard to eliminate
all noises.
 Next problem was the long cable connecting the microphone with audio interface.
This analogue cable is long and runs parallel to a~bundle of high-current power cables, which
induces high level of non-uniform noise.
 Another problem was a~low dynamic range of the device. 
When the amplification was set too low, commands spoken from the bigger distance were ignored.
Too high amplification, on the other hand, makes the noises even worse and effects with signal clipping
when the user is near the robot.

\begin{figure}[htb]
	\includegraphics[width=\columnwidth]{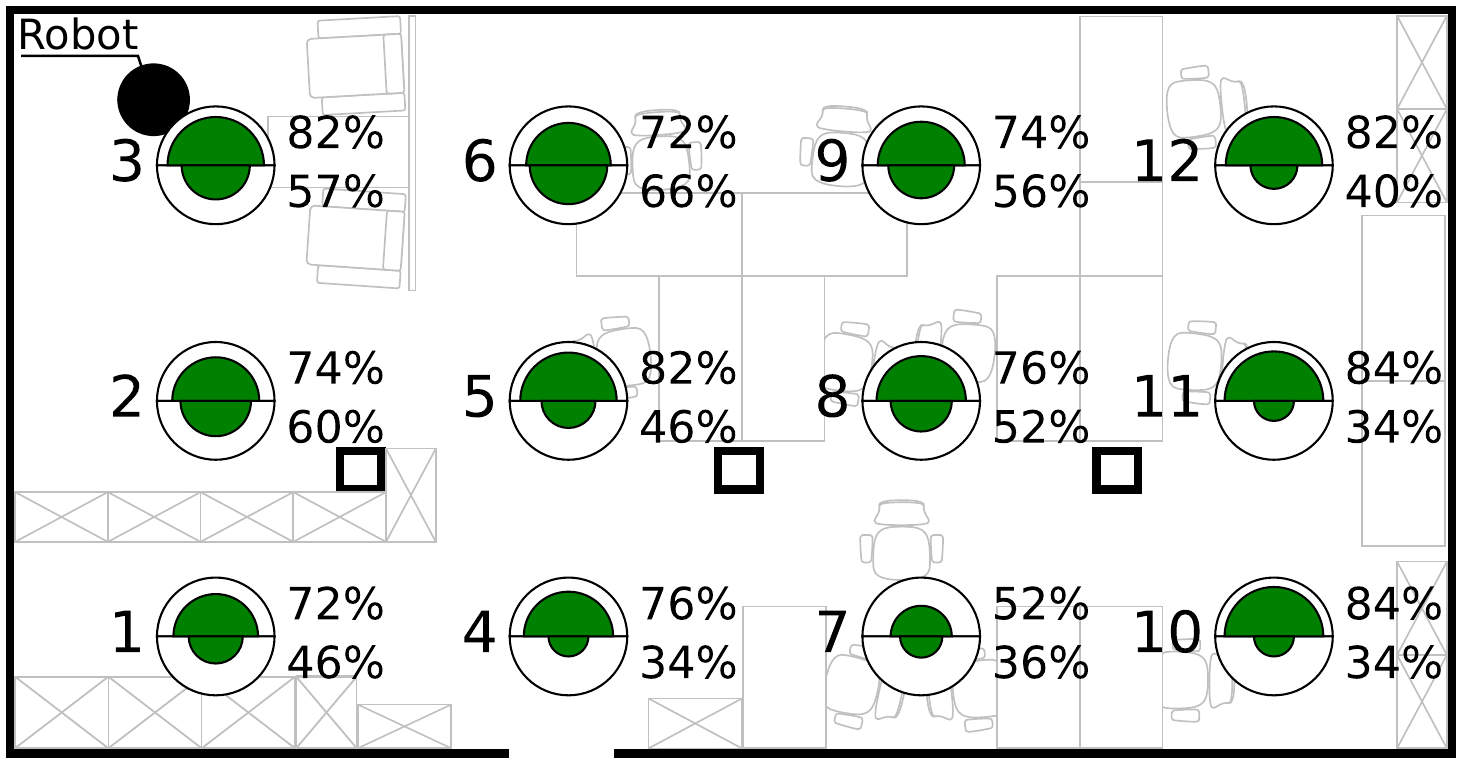}
	\caption{Comparison of keyword spotting accuracy for two microphones. 
		Top half represents omnidirectional microphone, bottom is built-in device.
		Position of the robot is depicted with black dot (near the spot 3)}
	\label{fig:voice-recognition-hm}
\end{figure}

To cope with those problems an additional microphone was added to the system. It is 
an omnidirectional device, with digital interface and high dynamic range. To mitigate mechanical
nosies, it was mounted using rubber bands. The recordings are much
cleaner and the overall recognition accuracy went up. There was, however, another problem still
occurring. Our testing environment is rather big (almost a~100 $m^2$ single room). There are also
some places, where the high closets stand on the line-of-sight between the robot and the operator.
In order to analyse the effect of the room shape and existing obstacles on the audio recognition accuracy 
word-spotting accuracy test was carried. During the experiment, the robot was static and different
operators were trying to trigger the robot using the keyword from 12 different places
(\Fig{fig:voice-recognition-hm}). Each of the 5 users said the keyword 10 times in each spot.
Keyword spotting was running on both internal and external microphone to compare the results.

The first visible thing is the much higher overall accuracy
of an external microphone. Not only the
results are better in every spot, but also they decay less with the distance (see the movie\footnote{\label{foot:video}\url{https://vimeo.com/410844697}}). In some spots
the accuracy dropped significantly. For example, the kitchen (spot $1$) is separated from the robot 
with the closets. Structure elements (pillars) also can interfere direct path from the operator
to the robot (spot~$7$). As the effective robot behaviours triggering is crucial, adding alternative
resources for this purpose was a~natural decision. It was achieved by utilising smart-home infrastructure
and expanding it with a~buttons. That kind of arrangement guarantees successful human-robot interaction.

\section{Conclusions}
\label{sec:conclusions}

The recent problems in our world let to the conclusion that high effort should be put to assistive robots research not only because of the ageing population of the developed countries. The number of worldwide diseases showed a~need to help people that have to be isolated or simultaneously are temporarily or continuously disabled. The robots can radically improve the situation even with simple applications like transportation or guarding. Such robots should communicate with patients in various ways to achieve reliable service of e.g. people that can not operate with hands or have problems with speaking. An approach proposed in this work, helps to develop control systems with the above assumptions. Currently, it used in the \tiago{} robot based system to help the Elderly in the various scenarios that include transportation attendance, guarding, fall prevention and hazard detection~\cite{incare-www-wut}.

%
%
\bibliographystyle{IEEEtran}
\bibliography{intent}
	
\end{document}